\documentclass[11pt]{article}
\usepackage{amsfonts}

\usepackage[final]{acl}

\usepackage{times}
\usepackage{latexsym}

\usepackage[T1]{fontenc}

\usepackage[utf8]{inputenc}

\usepackage{microtype}

\usepackage{inconsolata}

\usepackage{graphicx}

\usepackage{tcolorbox}
\tcbuselibrary{listings,breakable}
\usepackage{multirow}
\usepackage{array}
\newcolumntype{C}[1]{>{\centering\arraybackslash}p{#1}}
\usepackage{float}
\usepackage{booktabs}
\usepackage{tabularx}

\usepackage{amsmath}

\title{UKP\_Psycontrol at SemEval-2026 Task~2: Modeling Valence and Arousal Dynamics from Text}

\author{
 \textbf{Darya Hryhoryeva\textsuperscript{1,2}}
 \textbf{  Amaia Zurinaga\textsuperscript{3}}
 \textbf{  Hamidreza Jamalabadi\textsuperscript{3}}
 \textbf{  Iryna Gurevych\textsuperscript{1,2}}\\
 \textsuperscript{1}Ubiquitous Knowledge Processing Lab (UKP Lab), Technical University of Darmstadt \\
  \textsuperscript{2}National Research Center for Applied Cybersecurity ATHENE, Germany \\
 \textsuperscript{3}Psychiatric Control Systems Lab, Marburg University
}

\begin{document}
\maketitle
\begin{abstract}
This paper presents our system developed for SemEval-2026 Task~2. The task requires modeling both current affect and short-term affective change in chronologically ordered user-generated texts. We explore three complementary approaches: (1) LLM prompting under user-aware and user-agnostic settings, (2) a pairwise Maximum Entropy (MaxEnt) model with Ising-style interactions for structured transition modeling, and (3) a lightweight neural regression model incorporating recent affective trajectories and trainable user embeddings. Our findings indicate that LLMs effectively capture static affective signals from text, whereas short-term affective variation in this dataset is more strongly explained by recent numeric state trajectories than by textual semantics. Our system ranked first among participating teams in both Subtask~1 and Subtask~2A based on the official evaluation metric. \footnote{\url{https://github.com/UKPLab/semeval26_valence_arousal_from_text.git}}
\end{abstract}

\section{Introduction}

Emotions shape decision-making, well-being, and social interaction \cite{hofmann2014interpersonal}. NLP plays a central role in modeling affect at scale, as language provides a primary medium through which emotional states are expressed and inferred. In computational affect analysis, emotions are often represented using the affective circumplex \citep{russell1980circumplex}, which models affect along continuous dimensions of valence and arousal. Most NLP research relies on social media or review data \citep{zhang2026affective}, where affect is annotated by external raters or approximated via sentiment proxies. Such approaches capture expressed or perceived affect, offering only indirect access to internal emotional states \citep{buechel-hahn-2017-emobank}.

\textbf{SemEval 2026 Task~2: Predicting Variation in Emotional Valence and Arousal over Time from Ecological Essays} \citep{soni-etal-2026-semeval} addresses this limitation with a longitudinal dataset of self-reported affect. It contains chronologically ordered essays and feeling words written by U.S. service-industry workers over several years, each paired with self-assessed valence and arousal ratings. The task focuses on modeling subjectively experienced emotion from first-person accounts.

The shared task includes three subtasks: Subtask~1 (affect assessment per text) and Subtasks~2A and 2B (forecasting future affect). We propose solutions for Subtask~1 and Subtask~2A using: (1) LLM-based prompting, leveraging evidence that LLMs correlate strongly with human valence and arousal ratings \citep{Martinez2024UsingLL, broekens-affective-llm}; (2) a pairwise Maximum Entropy (MaxEnt) model with Ising interactions, motivated by the assumption that mental states evolve on an underlying energy landscape \cite{stocker2023formalizing,kheirkhah2025re} and follow a Boltzmann distribution \citep{jamalabadi2022complex,teutenberg2025synergistic};  and (3) a neural regression model using short-term affect trajectories and user embeddings.

Our system ranks first overall in both subtasks according to the official shared task leaderboard.
In Subtask~1 (25 teams), it achieves the best arousal and sixth-best valence performance; 
in Subtask~2A (14 teams), it ranks first overall, with top arousal and second-best valence results.

\section{Background}

The SemEval task models affect over time in text.
Subtask~1 (Longitudinal Affect Assessment) requires predicting valence and arousal for each text in a chronological sequence, for seen and unseen users. Subtask~2A (Forecasting Future Variation) predicts next-step changes in valence and arousal from prior texts and affective states. All experiments use the official SemEval evaluation interface \citep{semeval2026eval}.

Subtask~1 is evaluated with Pearson correlation ($r$) and MAE in between-user, within-user, and composite (combining both via Fisher $z$) settings. Subtask~2A reports user-level Pearson correlation and MAE for valence and arousal. All reported results follow the official scoring protocol.

The organizers provide L2-regularized ridge regression baselines with BERT-base-uncased \citep{devlin-etal-2019-bert} embeddings. We refer to these models using the shorthand names shown in parentheses, which are used consistently in the Results section. For Subtask~1, \textit{linear(BERT)} predicts affect from averaged token embeddings. For Subtask~2A, models predict state change ($\Delta_i = y_{i+1} - y_i$) using the current embedding with the previous score (\textit{linear(BERT; prev)}) or the previous score alone (\textit{linear(prev)}).

\section{Data}

The training set contains 2,764 entries from 137 users. Each includes a response to “How are you feeling?” and a self-reported valence (0--4) and arousal (0--2) from an affective circumplex grid (e.g., “Active” $\rightarrow$ 2,1). 
Texts are free-form essays (52\%) or lists of feeling words (48\%). 
Users contributed 20 entries on average (median: 14). Data span seven two-week periods, with 92\% of users participating in only one period. Common quality issues include invariant valence/arousal across entries and low-content or repetitive texts.

\section{System Overview}

We address the subtasks with three approaches. For Subtask~1, we use a MaxEnt model with pairwise interactions and an LLM-based method. For Subtask~2A, we additionally apply a neural regression model to predict continuous emotion values.

\subsection{LLM-based Prompting}
We leverage two LLMs: GPT-OSS 120B~\citep{openai2025gptoss120bgptoss20bmodel} for experiments and comparisons, and GPT-5~\citep{openai2025gpt5systemcard} for final submissions. We evaluate the effect of varying the number of shots under multiple conditions \citep{brown2020language}.

We distinguish between prompts that ignore or incorporate user-specific information. \textbf{User-agnostic} (Appendix~\ref{sec:appendix_agnostic}): Prompts include label-balanced random training examples from different users, and the model predicts emotions for batches of unseen texts. The validation set is split into roughly equal-sized batches. \textbf{User-aware} (Appendix~\ref{sec:appendix_aware}): Prompts contain chronologically ordered examples from one user, and the model predicts emotions for that user’s remaining texts. The prompt instructions encourage capturing individual expression patterns.

Using the affective circumplex grid from data collection, we compare two output formats. In one setting, the model generates\textbf{ textual emotion labels}, which are later mapped to numerical valence and arousal scores (Appendix~\ref{sec:appendix_mapping}). We evaluate this format against direct numerical prediction to assess which representation performs better. In the second setting, the model directly predicts \textbf{numerical scores}, either separately (two prompts) or jointly (single prompt). Prior work shows that LLMs tend toward overconfident in-context predictions \citep{li-etal-2025-large-language-models, wang-etal-2025-towards-objective}. In our preliminary experiments, we observed a bias toward stronger numerical labels. To mitigate this, we instruct the model to select the weakest valence/arousal value supported by the textual evidence.

We additionally predict emotions separately for \textbf{feeling words and essays}, which differ in length and contextual richness. We hypothesize that these differences may affect model behavior, such that separate prompting strategies yield more stable predictions.

We further implement a sliding-window in-context strategy, or \textbf{dynamic update}, using at most the last $N=15$ chronological examples per user. This allows conditioning on recent emotional expressions and capturing short-term dynamics. As new predictions are generated, they replace older examples, gradually filling the window with the predicted instances. For unseen users, the window is initialized with random examples and progressively replaced with the user’s own predictions, providing user-specific grounding. In both settings, essays and feeling words are processed separately within their respective sliding windows.

Across all experiments, the temperature parameter was fixed at 0.1 for GPT-OSS, whereas GPT-5 was used with its default API configuration.

\subsection{Semantic Binary Vectors}
\label{sec:semantic-binary-vec}
To support Subtask~2A, we represent each text as a low-dimensional semantic vector derived from emotion clusters. We first extract the 60 most frequent emotion-related words from the training data and use GPT-5 \citep{openai2025gpt5systemcard} to partition them into 10 semantically coherent clusters.
Next, each additional feeling word is assigned to exactly one of the predefined clusters. This allows all feeling-word examples to be represented as 10-dimensional binary vectors. For essays, we provide the LLM with the cluster names and representative frequent words and prompt it to assign the relevant clusters to each text. Each example is then encoded as a binary vector indicating the presence of the corresponding emotion clusters. All prompts are provided in Appendix~\ref{sec:appendix_binary}.

\subsection{MaxEnt with Ising}

We employ a MaxEnt model with Ising-style interactions to capture structured dependencies between affective states, semantic latent variables, and (when applicable) transitions. The same formulation is used for both Subtask~1 (state prediction) and Subtask~2A (transition prediction). Intuitively, the model can be viewed as a probabilistic graphical model over binary variables, where each component (e.g., valence level, arousal level, or semantic feature) contributes an individual bias term, while pairwise interaction terms capture how variables co-occur or influence each other. This allows the model to represent dependencies such as certain semantic patterns being associated with specific affective configurations or transitions.

The model defines a probability distribution over binary state vectors $\mathbf{x}$:

\begin{equation}
P(x) = \frac{1}{Z} \exp(-E(x)),
\end{equation}

with energy

\begin{equation}
E(x) = - \mathbf{x}^\top \mathbf{h}
      - \frac{1}{2} \mathbf{x}^\top \mathbf{J} \mathbf{x},
\end{equation}

where $\mathbf{h}$ models linear effects and $\mathbf{J}$ captures pairwise interactions. Since the affective state space is discrete and bounded, we explicitly enumerate all valid state–transition combinations and compute the partition function $Z$ exactly, enabling maximum likelihood training.

All categorical affective variables are transformed via one-hot encoding (valence, arousal, and, for Subtask~2A, $\Delta v$ and $\Delta a$). To incorporate semantic information while keeping exact normalization tractable, semantic binary LLM-derived vectors are compressed using an autoencoder (Appendix~\ref{sec:appendix_autoencoder}). The resulting latent vectors are binarized and appended to the state vector. These semantic variables act as additional nodes in the model, allowing the MaxEnt formulation to learn interactions between affective states and higher-level semantic cues. In practice, they provide complementary information to the affective variables and contribute to performance improvements.

At inference time, the model estimates the joint distribution over affective states and latent variables. For Subtask~1, we compute the conditional distribution over valence and arousal and decode predictions via conditional expectations, yielding continuous estimates aligned with correlation-based evaluation. For Subtask~2A, we marginalize latent variables to obtain $P(\Delta v, \Delta a \mid v_t, a_t)$ and compute expectation-based transitions. This strategy leverages the full learned distribution and produces smooth, calibrated outputs for both tasks.

\subsection{Neural Regression Model}
\label{sec:regression}

For Subtask~2A, we train a lightweight neural regression model to predict next-step changes in valence and arousal using a sliding window (1–4 prior entries) per user.

At time $t$, inputs include embeddings of recent texts, current valence and arousal, the previous state change, and a trainable user embedding capturing individual dynamics. Texts are encoded with RoBERTa-base \citep{liu2019robertarobustlyoptimizedbert} and mean-pooled.

We compare three input settings: (1) a \textbf{no-text baseline} using only affective features and user embeddings; (2) a \textbf{text-enhanced model} with contextualized embeddings; and (3) \textbf{semantic cluster representations}, embedding emotion cluster names instead of binary indicators (Section~\ref{sec:semantic-binary-vec}).

\section{Results}

\begin{table*}
\centering
\small
\setlength{\tabcolsep}{4pt}
\renewcommand{\arraystretch}{1.1}

\begin{tabular}{lllcccc}
\hline
& &
& \multicolumn{2}{c}{\textbf{Valence}} 
& \multicolumn{2}{c}{\textbf{Arousal}} \\

\textbf{Dataset} & \textbf{Approach} & \textbf{System}
& \textbf{$r$ composite} 
& \textbf{MAE composite} 
& \textbf{$r$ composite} 
& \textbf{MAE composite} \\
\hline
\hline

\multirow{16}{*}{dev}
& \multirow{8}{*}{\parbox{2cm}{LLM-based \\ user-aware}}
& emo 10 & 0.617 & 0.655 & \textbf{0.380} & 0.431 \\ 
& & emo 15 & 0.647 & 0.590 & 0.370 & 0.420 \\ 
& & emo 20 & \textbf{0.661} & 0.568 & 0.378 & 0.433 \\ 
& & emo w/e split 15 & 0.616 & 0.642 & 0.334 & 0.421 \\ 
& & v/a split 15 & 0.640 & 0.617 & 0.368 & 0.431 \\ 
& & v/a joint 15 & 0.643 & 0.572 & 0.292 & 0.495 \\ 
& & v/a split w/e split 15 & 0.642 & 0.648 & 0.364 & 0.426 \\ 
& & emo dynamic 15 & 0.637 & 0.635 & 0.365 & 0.437 \\ 

\cline{2-7}

& \multirow{5}{*}{\parbox{2cm}{LLM-based \\ user-agnostic}}
& emo 15 & 0.629 & 0.685 & 0.351 & 0.417 \\
& & emo granular 15 & \textbf{0.657}  &  0.701  &  \textbf{0.374}  &  0.411 \\
& & emo w/e split 15 & 0.647  &  0.684  &  0.364  &  0.437 \\
& & emo dynamic 15 & 0.629 & 0.713 & 0.356 & 0.435 \\
\cline{3-7}
& & emo 15 GPT-5 & 0.670  &  0.641  &  0.359  &  0.425 \\

\cline{2-7}

& \parbox{2cm}{MaxEnt}
& Ising (expectation) & 0.533 & 0.831 & 0.277 & 0.478 \\

\hline

\multirow{3}{*}{test}
& \multirow{1}{*}{\parbox{2cm}{LLM-based}}
& submission & 0.667 & 0.595 & 0.554 & 0.345 \\

\cline{2-7}

& \parbox{2cm}{MaxEnt}
& Ising & 0.589 & 0.701 & 0.327 & 0.448 \\

\cline{2-7}

& \multirow{1}{*}{\parbox{2cm}{Baseline}}
& linear(BERT) & 0.557 & 0.743 & 0.299 & 0.459 \\

\hline

\end{tabular}
\caption{Subtask 1: Comparative performance of our systems and baselines across models and prompting settings. Best results within the user-aware and user-agnostic approaches are bolded. The official submission uses GPT-5 and combines the best-performing configurations.}
\label{tab:results-subtask1}
\end{table*}

LLM-based systems produced deterministic predictions, whereas the MaxEnt and neural regression models generated continuous estimates.

\subsection{Longitudinal Affect Assessment}
Table~\ref{tab:results-subtask1} reports the results for Subtask~1 on the development and test sets.

\paragraph{LLM-based solutions}
\textbf{User-aware} prompting performs marginally better than the \textbf{user-agnostic} variant. This trend is expected, as the user-aware setup provides user-specific examples, enabling adaptation to individual expression patterns. 
The performance gap, however, remains small, indicating that label-balanced random demonstrations approximate much of the benefit of explicit user history.
Qualitative comparisons between the user-aware (\textit{emo 15}) and user-agnostic (\textit{emo granular 15}) configurations are presented in Appendix~\ref{sec:appendix_examples_llm}.

Increasing the \textbf{number of shots} improves valence correlation in the user-aware setting (\textit{emo 10}: 0.617; \textit{emo 15}: 0.647; \textit{emo 20}: 0.661), while arousal does not exhibit a comparable trend. 
This suggests valence benefits more from additional context.

Direct \textbf{numerical prediction} (\textit{v/a split 15}, \textit{va joint 15}) does not outperform textual emotion prediction (\textit{emo 15}) followed by label-to-score mapping, suggesting that natural language emotion descriptors are better aligned with LLM pretraining and yield more stable predictions than direct numeric elicitation.

Contrary to our initial hypothesis, \textbf{separating essays and feeling words} into distinct prompts does not consistently improve performance over mixed-input prompting.   The performance drop is more pronounced in the user-aware setting (\textit{emo w/e split 15} vs. \textit{emo 15}), consistent with the disruption of chronological user history.

The \textbf{dynamic update} strategy underperforms fixed-shot prompting in both settings (\textit{emo dynamic 15} vs. \textit{emo granular 15}). Since the sliding window gradually incorporates model predictions, errors may accumulate and propagate. This effect may introduce compounding noise into the prompt context, leading to reduced stability compared to fixed-shot prompting based solely on gold examples.

A representative experiment with \textit{GPT-5} (\textit{emo 15 GPT-5}) shows improved performance over GPT-OSS 120B. All ablations were conducted with GPT-OSS 120B under the assumption that relative differences transfer to stronger models, while final submissions were generated using GPT-5.

\paragraph{MaxEnt model}

The MaxEnt model with Ising-style latent interactions provides a structured probabilistic approach, estimating a joint distribution over binary latent semantic factors and affective outcomes. Expectation-based decoding produces continuous predictions aligned with the evaluation metric.

In our experiments, the MaxEnt system performs below LLM-based models. This may reflect limitations of the binarized semantic representation, which can reduce affective nuance, and the absence of explicit temporal modeling for longitudinal prediction. 
The approach offers a transparent probabilistic alternative with an interpretable latent structure, though less expressive than LLMs.

\begin{table*}
\centering
\small
\setlength{\tabcolsep}{4pt}
\renewcommand{\arraystretch}{1.1}

\begin{tabular}{lll C{1.2cm} C{1.2cm} C{1.2cm} C{1.2cm}}
\hline
& &
& \multicolumn{2}{c}{\textbf{Valence}} 
& \multicolumn{2}{c}{\textbf{Arousal}} \\

\textbf{Dataset} & \textbf{Approach} & \textbf{System}
& \textbf{$r$} 
& \textbf{MAE} 
& \textbf{$r$} 
& \textbf{MAE} \\
\hline
\hline

\multirow{7}{*}{dev}
& \multirow{2}{*}{\parbox{2cm}{LLM-based}}
& emo 20 & 0.346 & 1.058 & 0.525 & 0.679 \\ 
& & v/a split 20 & 0.469 & 0.934 & 0.607 & 0.628 \\ 

\cline{2-7}

& \multirow{2}{*}{\parbox{2cm}{Regression}}
& best valence & \textbf{0.626} & 0.873 & -- & -- \\ 
& & best arousal & -- & -- & \textbf{0.654} & 0.632 \\ 

\cline{2-7}

& \parbox{2cm}{MaxEnt}
& Ising (expectation) & 0.526 & 1.143 & 0.651 & 0.615 \\
\cline{2-7}

& \multirow{2}{*}{\parbox{2cm}{Baseline}}
& linear(prev) & 0.520 & 0.987 & 0.609 & 0.635 \\ 
& & linear(BERT; prev) & 0.305 & 1.192 & 0.408 & 0.774 \\ 

\hline

\multirow{4}{*}{test}
& \multirow{1}{*}{\parbox{2cm}{Regression}}
& submission & 0.675 & 1.118 & 0.683 & 0.689 \\

\cline{2-7}

& \parbox{2cm}{MaxEnt}
& Ising & 0.615 & 1.196 & 0.670 & 0.652 \\
\cline{2-7}

& \multirow{2}{*}{\parbox{2cm}{Baseline}}
& linear(prev) & 0.615 & 1.168 & 0.670 & 0.638 \\
& & linear(BERT; prev) & 0.430 & 1.251 & 0.405 & 0.708 \\

\hline

\end{tabular}
\caption{Subtask~2A: Comparative performance of our systems and baselines. Best development results are bolded. The official submission combines the best-performing configurations.}
\label{tab:results-subtask2a}
\end{table*}

\paragraph{Final submission system}

For the official submission (\textit{submission}), we combined the strongest user-aware (\textit{emo 15}) and user-agnostic (\textit{emo granular 15}) configurations based on textual emotion labels and ran them with GPT-5. For seen users, we used chronologically grouped batches from two-week periods; for unseen users, we employed randomly sampled granular batches.

On the test set, the system achieves $r=0.667$ for valence and $r=0.554$ for arousal, outperforming the \textit{linear(BERT)} baseline. The improvement is particularly pronounced for arousal. Detailed metrics indicate that the strong arousal performance is largely driven by seen users, likely due to the higher number of available in-context examples at test time.

Additionally, performance on essays-only subsets (valence: 0.685; arousal: 0.500 in detailed test metrics) confirms that LLMs are especially effective at extracting affective signals from richer free-text inputs.

\begin{table}
  \centering
  \begin{tabular}{lcc}
    \hline
    \textbf{Parameters} & \textbf{Best valence} & \textbf{Best arousal} \\
    \hline
    Target     & Both & Arousal  \\
    History length     & 2 & 1  \\
    Text     & No & No \\
    Sem clusters     & No & No  \\
    User emb dim     & 2  & 4 \\
    \hline
  \end{tabular}
  \caption{Best system configurations for Subtask~2A.}
  \label{tab:subtask2a-best}
\end{table}

\subsection{Forecasting Variation in Affect}

Table~\ref{tab:results-subtask2a} presents the results for Subtask~2A on the development and test sets. We also include baseline results on the development set, following the organizers’ published description for direct comparability.

\paragraph{LLM-based solutions}
We adapted the prompting strategies from Subtask~1 to predict affective change using either textual emotion labels (\textit{emo 20}) or direct numerical prediction (\textit{v/a split 20}). On the development set, neither variant outperforms the \textit{linear(prev)} baseline. However, direct numerical prediction consistently performs better than label-based prompting, suggesting that forecasting affective change benefits from scalar modeling rather than categorical emotion representations.

\paragraph{MaxEnt model}
The MaxEnt model (\textit{Ising}) shows competitive development performance, especially for arousal. Under expectation-based decoding, it surpasses the \textit{linear(prev)} baseline on arousal and approaches it in valence. These results suggest that jointly modeling states and transitions captures meaningful dependencies, particularly for short-term arousal shifts.

For valence, performance remains competitive but below the best regression setup, which is consistent with greater variability or longer-range dependencies not captured by conditioning on the current state alone. MaxEnt models with pairwise interactions are sensitive to limited sample sizes, as parameter growth can introduce estimation bias \citep{Macke2011}. Given the small dataset and quadratic increase in interaction parameters, such effects may constrain gains over simpler regression methods.

A limitation of the current formulation is that it only captures instantaneous (pairwise) interactions and does not explicitly model temporal dependencies beyond adjacent states. Extending the MaxEnt framework to incorporate temporal interactions (e.g., by introducing time-lagged couplings or higher-order dynamics) could enable better modeling of affective trajectories. However, such extensions would further increase the number of parameters and may require stronger regularization or more data to be effective.

Nevertheless, the framework shows that structured probabilistic modeling over a constrained state space is a viable alternative for forecasting affective change.

\paragraph{Neural regression model}

The neural regression model yields the strongest development performance (see Appendix~\ref{sec:appendix_subtask2a_res} for full results). Shorter history windows outperform longer ones, suggesting that recent affect is most predictive of immediate change.
Models without textual or semantic cluster features perform best: using only previous valence, arousal, state change, and user embeddings gives the highest scores. 
This indicates that numeric trajectories provide a cleaner forecasting signal than text features.
We also observe asymmetry across targets: arousal performs better in single-task training, while valence benefits from joint prediction. Arousal may exhibit stronger short-term autocorrelation and depend mainly on its own recent trajectory, whereas valence benefits from shared representations due to its interaction with arousal, making joint training more helpful.

\paragraph{Final submission system}
For the official submission, we combined the best-performing regression configurations for valence and arousal (Table~\ref{tab:subtask2a-best}) and retrained on the full dataset, including the development set. On the test set, the final system outperforms both baselines for valence and arousal, with consistent gains over the \textit{linear(prev)} baseline, highlighting the benefit of user representations and short-term affective trajectories.

\section{Conclusion}
We present three modeling approaches for longitudinal affect assessment and short-term forecasting in SemEval-2026 Task~2. Our experiments show that LLM-based prompting is highly effective for static valence and arousal prediction, especially when user-specific context is incorporated. In the small-sample setting of this task, LLM prompting clearly outperformed probabilistic models, suggesting that linguistic representations encode a relatively consistent structure that maps free text to core affective dimensions such as valence and arousal.

In contrast, short-term forecasting of affective change benefited more from structured neural regression based on recent affect trajectories and probabilistic transition modeling, while textual representations provided only limited additional predictive gains for temporal dynamics.

\section*{Limitations}
Emotional well-being dimensions are typically characterized by slow temporal dynamics, with most spectral power occurring over longer time scales (e.g., weeks) rather than day-to-day fluctuations \cite{jamalabadi2025optimizing}. In this context, it remains unclear whether our findings would generalize to datasets spanning longer observation periods, especially given that 92\% of subjects in this shared task have data restricted to a single two-week interval. Performance patterns may differ when modeling longer trajectories that better capture the slow dynamics of mental disorder progression.

Furthermore, the test set for Subtask~2A is relatively small (46 instances), and we observed substantial variability in evaluation scores under minor data perturbations and across modeling choices. This suggests that small leaderboard differences should be interpreted with caution.

LLM-based prompting showed sensitivity to temperature: higher temperatures occasionally yielded improved performance but also increased variance across runs, indicating limited stability. 

In addition, our reliance on proprietary LLMs raises concerns regarding reproducibility, privacy, and data governance when processing sensitive self-reported emotional content. Future work could investigate privacy-aware modeling frameworks to improve reproducibility and ensure responsible handling of sensitive emotional data.

\section{Acknowledgements}
This work was supported by the DYNAMIC center, which is funded by the LOEWE program of the Hessian Ministry of Science and Arts (Grant Number: LOEWE/1/16/519/03/09.001(0009)/98).

This research was further funded by the German Federal Ministry of Research, Technology and Space and the Hessian Ministry of Higher Education, Research, Science and the Arts within their joint support of the National Research Center for Applied Cybersecurity ATHENE.

Many thanks to Bhavyajeet Singh for the thoughtful discussions.

\bibliography{custom}

\appendix
\section{Textual emotions to numerical values mapping}
\label{sec:appendix_mapping}

Table~\ref{tab:emotion-va-mapping} shows the discrete mapping from textual emotion labels to numerical valence and arousal levels used in our experiments.

\begin{table}[H]
  \centering
  \begin{tabular}{lcc}
    \hline
    \textbf{Emotion label} & \textbf{Valence} & \textbf{Arousal} \\
    \hline
    Jittery, nervous     & 0 & 2  \\
    Somewhat jittery     & 1 & 2  \\
    Active     & 2 & 2 \\
    Somewhat lively     & 3 & 2  \\
    Lively, enthusiastic      & 4  & 2 \\
    Very sad     & 0 & 1 \\
    Somewhat sad & 1 & 1 \\
    Neutral & 2 & 1 \\
    Somewhat happy & 3 & 1 \\
    Very happy & 4 & 1  \\
    Sluggish, tired & 0 & 0 \\
    Somewhat sluggish & 1 & 0 \\
    Quiet & 2 & 0 \\
    Somewhat content & 3 & 0 \\
    Content, calm & 4 & 0 \\
    \hline
  \end{tabular}
  \caption{Textual emotion label to numerical valence and arousal mapping.}
  \label{tab:emotion-va-mapping}
\end{table}

\section{Semantic binary vector generation prompts}
\label{sec:appendix_binary}

In this section, we present the prompts we used to generate semantic binary vectors.

\subsection{Split most frequent feeling words into clusters}
\label{sec:appendix_binary_1}

\begin{tcolorbox}[breakable, colback=white, colframe=black]
{\footnotesize\ttfamily
You are an expert in emotions and feelings. Given the following list of words and phrases, divide them into exactly 10 categories. Each category should group words and phrases that express similar emotions, and each category must have a clear, descriptive name.\\
- Preserve the original order of the words within each category as they appear in the input list. Check this requirement programmatically.\\
- Each word or phrase must appear exactly once -- no duplicates and none omitted.\\
- Present the result in a plain Python-friendly dictionary format, without any explanations or comments.\\ 
Output format example: ...\\
List: ...
}
\end{tcolorbox}

\subsection{Split all feeling words into clusters}
\label{sec:appendix_binary_2}

\begin{tcolorbox}[breakable, colback=white, colframe=black]
{\footnotesize\ttfamily
You are an expert in emotions and feelings. Below are 10 predefined emotion categories, each grouping words and phrases that express similar emotions and labeled with a descriptive name. Your task is to assign each word or phrase from the provided list to exactly one of these categories -- the one that best describes its meaning.\\
If a word or phrase does not explicitly describe an emotion, assign it based on the emotional state implied when a person uses it to describe how they feel.\\
Examples:\\
“not so happy” → Sad / Lonely\\
“dizzy” → Unwell / Physical State\\
“running behind” → Anxious / Worried\\
Instructions:\\
- Do not modify the category names or their contents.\\
- Preserve the original order of the input words within each category as they appear in the input list.\\
- Programmatically verify that every word or phrase from the input appears exactly once in the output - no duplicates and none omitted.\\
- Present the result in a plain Python-friendly dictionary format, without any explanations or comments.\\
- Treat every item in the list as-is -- including typos, fragments, and non-emotions -- and still force them into one of the categories.\\
Output format example: ...\\
Categories: ...\\
List: ...
}
\end{tcolorbox}

\subsection{Split all essays into clusters}
\label{sec:appendix_binary_3}

\begin{tcolorbox}[breakable, colback=white, colframe=black]
{\footnotesize\ttfamily
You are an expert in emotions and human feelings.
Below are 10 predefined emotion categories, each with example words that describe it.\\
Your task is: given a list of user essays in which a person describes their day, assign exactly from 3 to 4 emotion categories to each essay that best represent the person’s feelings.\\
Instructions:\\
 - Preserve the original order of the essays as they appear in the input.\\
 - Use predefined categories only.\\
 - Prioritize the emotional meaning conveyed in the essay over the presence of example words in the categories -- choose categories that best reflect the person’s feelings, even if exact keywords are not present.\\
 - Ignore typos, gibberish, or any content that is not meaningful. Do not attempt to analyze unrelated text.\\
 - If an essay is completely unrelated or does not describe a person’s day or feelings, mark it as “NOT\_RELEVANT”.\\
 - Present the output only in a Python dictionary format, without any explanations, comments, or extra text.\\
 - Each essay must appear exactly once in the output.\\
 - Do not ask for clarification under any circumstances.\\
Output format example: ... \\
Categories: ... \\
Essays: ...
}
\end{tcolorbox}

\section{User-aware prompt}
\label{sec:appendix_aware}

In this section, we provide the prompts we used to predict emotions for \textbf{seen} users.
\begin{tcolorbox}[breakable, colback=white, colframe=black]
{\footnotesize\ttfamily
You are an expert in human emotions. Below is a chronological sequence of short texts written by the same user, each describing how they felt on a particular day.
Your task is to assign exactly one emotion from the allowed list to each text — the emotion that best matches the overall feeling expressed.\\
The user has a personal, consistent way of expressing emotions. Learn from the previously labeled examples how this specific user tends to describe their emotional states, and apply this understanding when labeling the new texts. Your labels should follow the user’s own pattern of emotional expression, not generic interpretations. \\
Instructions:\\
- If the text does not explicitly describe feelings, choose the emotion that best fits the emotional state implied by the text.\\
- All texts in the evaluation set come from the same user as the examples.\\
- Preserve the original order of the texts.\\
- Each text must appear exactly once in the output — no duplicates and none omitted.\\
- Present the result in a plain Python-friendly dictionary format, without any explanations or comments.\\ 
Allowed emotions:\\
{{"Jittery, nervous", "Somewhat jittery", "Active", "Somewhat lively", "Lively, enthusiastic", "Very sad", "Somewhat sad", "Neutral", "Somewhat happy", "Very happy", "Sluggish, tired", "Somewhat sluggish", "Quiet", "Somewhat content", "Content, calm"}} \\
Previous texts with assigned emotions: ... \\
Sequence of texts for evaluation: ...
}
\end{tcolorbox}

\section{User-agnostic prompt}
\label{sec:appendix_agnostic}

In this section, we provide the prompts we used to predict emotions for \textbf{unseen} users.
\begin{tcolorbox}[breakable, colback=white, colframe=black]
{\footnotesize\ttfamily
You are an expert in human emotions. Below is a list of short texts, where each text describes how a person feels today. Your task is to assign exactly one emotion from the allowed list to each text -- the emotion that best matches the overall feeling expressed.\\ 
Instructions:\\
- If the text does not explicitly describe feelings, choose the emotion that best fits the emotional state implied by the text.\\
- Preserve the original order of the texts.\\
- Each text must appear exactly once in the output -- no duplicates and none omitted.\\
- Present the result in a plain Python-friendly dictionary format, without any explanations or comments. \\
Emotions: ...\\
Examples: ...\\
Output format example: ...\\
List of texts: ...
}
\end{tcolorbox}

\section{LLM-based prompting examples}
\label{sec:appendix_examples_llm}

Table~\ref{tab:examples-llm} illustrates qualitative differences between the user-aware and user-agnostic LLM-based predictions. Texts with IDs~1--2 and 3--4 form two pairs of consecutive entries from the same user, originating from different time periods.

The original annotations indicate that this user exhibits stable valence values (V=2) across entries. The user-aware setting successfully reproduces this individual consistency, whereas the user-agnostic predictions show larger variation in valence. This suggests that incorporating user history helps the model capture persistent affective tendencies beyond lexical cues.

For arousal, both approaches produce predictions close to the original annotations, indicating that arousal appears to be more directly inferable from local textual signals than valence in this example.

\begin{table*}
  \centering
  \begin{tabularx}{\textwidth}{lXcccccc}
    \toprule
    & & \multicolumn{2}{c}{\textbf{Original}}
      & \multicolumn{2}{c}{\textbf{User-aware}}
      & \multicolumn{2}{c}{\textbf{User-agnostic}} \\

    \cmidrule(lr){3-4}
    \cmidrule(lr){5-6}
    \cmidrule(lr){7-8}

    \textbf{ID} & \textbf{Text}
      & \multicolumn{1}{c}{\textbf{V}}
      & \multicolumn{1}{c}{\textbf{A}}
      & \multicolumn{1}{c}{\textbf{V}}
      & \multicolumn{1}{c}{\textbf{A}}
      & \multicolumn{1}{c}{\textbf{V}}
      & \multicolumn{1}{c}{\textbf{A}} \\
    \midrule

    1 & Happy, Respected, Hopeful, Energy, Active
      & \multicolumn{1}{c}{2} & \multicolumn{1}{c}{2}
      & \multicolumn{1}{c}{2} & \multicolumn{1}{c}{2}
      & \multicolumn{1}{c}{4} & \multicolumn{1}{c}{1} \\

    2 & Happy, Excited, Motivation, Energy, Mindful
      & \multicolumn{1}{c}{2} & \multicolumn{1}{c}{2}
      & \multicolumn{1}{c}{2} & \multicolumn{1}{c}{2}
      & \multicolumn{1}{c}{4} & \multicolumn{1}{c}{2} \\

    3 & I got dressed and ready for my run. I am currently having breakfast now. I think after breakfast and my run I will do some decor house shopping. I like to go to thrift stores also it's just a long day digging to find goods.
      & \multicolumn{1}{c}{2} & \multicolumn{1}{c}{2}
      & \multicolumn{1}{c}{2} & \multicolumn{1}{c}{2}
      & \multicolumn{1}{c}{2} & \multicolumn{1}{c}{2} \\

    4 & I'm feeling energetic. I went to the gym. I did a couple miles this morning. I organized the bathroom closet. Now I am getting ready to prepare dinner. It's also Friday so that makes me happy. Not sure what I am doing tonight yet. It's also raining.
      & \multicolumn{1}{c}{2} & \multicolumn{1}{c}{2}
      & \multicolumn{1}{c}{2} & \multicolumn{1}{c}{2}
      & \multicolumn{1}{c}{4} & \multicolumn{1}{c}{2} \\

    \bottomrule
  \end{tabularx}
  \caption{\label{tab:examples-llm}
    Qualitative comparison of original annotations and LLM predictions on the development set. V denotes valence, and A denotes arousal. Texts 1 and 2 are examples of feeling-word lists, texts 3 and 4 are free-form essays.
    Qualitative comparison of original annotations and LLM predictions on the development set. V denotes valence, and A denotes arousal. Texts 1 and 2 are examples of feeling-word lists, texts 3 and 4 are free-form essays.
  }
\end{table*}

\section{Autoencoder for semantic binary compression}
\label{sec:appendix_autoencoder}
To reduce the dimensionality of the binary semantic vectors derived from the LLM, we employ a lightweight fully connected autoencoder with a low-dimensional latent bottleneck. 
The encoder maps the 60-dimensional input vector to a 10-dimensional latent representation via a two-layer multilayer perceptron (60$\rightarrow$32$\rightarrow$10), the decoder mirrors this structure to reconstruct the original input.

The autoencoder is trained jointly with the MaxEnt model using a reconstruction objective, allowing the latent variables to retain task-relevant semantic information. 
The resulting latent representation is binarized and appended to the state vector, enabling the MaxEnt formulation to model interactions between affective variables and compressed semantic cues while keeping exact normalization computationally tractable.

\section{System setup and hyperparameters for neural regression}
\label{sec:appendix_hyperparameters}
The neural regression system experiments for Subtask~2A were conducted on a single NVIDIA T4 GPU with a batch size of 32. We optimized the model using AdamW \citep{loshchilov2019decoupledweightdecayregularization} with a learning rate of 5e-4 and mean squared error as the training objective. We used a 90/10 train-validation split and employed early stopping with a patience of 5 epochs.

The neural regression model is a feed-forward neural network. The concatenated input vector (text representation, affective features, previous delta, and user embedding) is passed through two fully connected layers with Layer Normalization, ReLU activations, and dropout regularization, followed by a linear output layer.

\section{Neural regression results for Subtask~2A}
\label{sec:appendix_subtask2a_res}
This section presents the complete experimental results for Subtask~2A obtained with the neural regression model. We report performance across all evaluated configurations, including different feature sets and training settings (Table~\ref{tab:appendix-full-results-1}, Table~\ref{tab:appendix-full-results-2}, Table~\ref{tab:appendix-full-results-3}).

\begin{table*}
\centering
\scriptsize
\setlength{\tabcolsep}{3pt}
\renewcommand{\arraystretch}{1.1}

\begin{tabular}{l c c c c  C{0.9cm} C{0.9cm}  C{0.9cm} C{0.9cm}}
\toprule
& & & &
& \multicolumn{2}{c}{\textbf{Valence}} 
& \multicolumn{2}{c}{\textbf{Arousal}} \\
\textbf{Target} & \textbf{History length} & \textbf{Text} & \textbf{Semantic cluster} & \textbf{User emb}
& \textbf{{$r$}} & \textbf{MAE}
& \textbf{{$r$}} & \textbf{MAE} \\
\midrule
both & 1 & No & No & 8 & 0.592 & 0.902 & 0.618 & 0.620 \\
both & 2 & No & No & 8 & 0.625 & 0.866 & 0.595 & 0.622 \\
both & 3 & No & No & 8 & 0.589 & 0.891 & 0.607 & 0.619 \\
both & 4 & No & No & 8 & 0.589 & 0.891 & 0.607 & 0.619 \\
\midrule
both & 1 & No & No & 2 & 0.531 & 0.975 & 0.582 & 0.659 \\
both & 2 & No & No & 2 & 0.626 & 0.873 & 0.627 & 0.620 \\
both & 1 & No & No & 4 & 0.615 & 0.890 & 0.570 & 0.643 \\
both & 2 & No & No & 4 & 0.568 & 0.911 & 0.605 & 0.620 \\
both & 1 & No & No & 16 & 0.595 & 0.876 & 0.570 & 0.627 \\
both & 2 & No & No & 16 & 0.578 & 0.910 & 0.566 & 0.627 \\
both & 1 & No & No & 32 & 0.603 & 0.900 & 0.611 & 0.600 \\
both & 2 & No & No & 32 & 0.589 & 0.890 & 0.591 & 0.610 \\
\midrule
arousal & 1 & No & No & 2 & -- & -- & 0.580 & 0.649 \\
arousal & 2 & No & No & 2 & -- & -- & 0.604 & 0.610 \\
arousal & 1 & No & No & 4 & -- & -- & 0.654 & 0.632 \\
arousal & 2 & No & No & 4 & -- & -- & 0.589 & 0.622 \\
arousal & 1 & No & No & 8 & -- & -- & 0.596 & 0.624 \\
arousal & 2 & No & No & 8 & -- & -- & 0.646 & 0.581 \\
arousal & 1 & No & No & 16 & -- & -- & 0.601 & 0.625 \\
arousal & 2 & No & No & 16 & -- & -- & 0.609 & 0.611 \\
arousal & 1 & No & No & 32 & -- & -- & 0.584 & 0.618 \\
arousal & 2 & No & No & 32 & -- & -- & 0.597 & 0.603 \\
\midrule
valence & 1 & No & No & 2 & 0.535 & 0.954 & -- & -- \\
valence & 2 & No & No & 2 & 0.616 & 0.893 & -- & -- \\
valence & 1 & No & No & 4 & 0.534 & 0.968 & -- & -- \\
valence & 2 & No & No & 4 & 0.588 & 0.901 & -- & -- \\
valence & 1 & No & No & 8 & 0.579 & 0.933 & -- & -- \\
valence & 2 & No & No & 8 & 0.595 & 0.916 & -- & -- \\
valence & 1 & No & No & 16 & 0.567 & 0.881 & -- & -- \\
valence & 2 & No & No & 16 & 0.604 & 0.871 & -- & -- \\
valence & 1 & No & No & 32 & 0.567 & 0.924 & -- & -- \\
valence & 2 & No & No & 32 & 0.609 & 0.873 & -- & -- \\
\bottomrule
\end{tabular}

\caption{Subtask~2A full experimental results with numeric affective features only.}
\label{tab:appendix-full-results-1}
\end{table*}

\begin{table*}
\centering
\scriptsize
\setlength{\tabcolsep}{3pt}
\renewcommand{\arraystretch}{1.1}

\begin{tabular}{l c c c c  C{0.9cm} C{0.9cm}  C{0.9cm} C{0.9cm}}
\toprule
& & & &
& \multicolumn{2}{c}{\textbf{Valence}} 
& \multicolumn{2}{c}{\textbf{Arousal}} \\
\textbf{Target} & \textbf{History length} & \textbf{Text} & \textbf{Semantic cluster} & \textbf{User emb}
& \textbf{{$r$}} & \textbf{MAE}
& \textbf{{$r$}} & \textbf{MAE} \\
\midrule
both & 1 & Yes & No & 8 & 0.554 & 0.920 & 0.596 & 0.616 \\
both & 2 & Yes & No & 8 & 0.560 & 0.901 & 0.603 & 0.611 \\
both & 3 & Yes & No & 8 & 0.570 & 0.922 & 0.606 & 0.627 \\
both & 1 & Yes & No & 16 & 0.590 & 0.874 & 0.554 & 0.639 \\
both & 2 & Yes & No & 16 & 0.598 & 0.885 & 0.603 & 0.614 \\
both & 3 & Yes & No & 16 & 0.619 & 0.847 & 0.575 & 0.626 \\
both & 1 & Yes & No & 32 & 0.621 & 0.855 & 0.580 & 0.625 \\
both & 2 & Yes & No & 32 & 0.606 & 0.885 & 0.566 & 0.629 \\
both & 3 & Yes & No & 32 & 0.617 & 0.845 & 0.599 & 0.610 \\
\midrule
arousal & 1 & Yes & No & 8 & -- & -- & 0.598 & 0.629 \\
arousal & 2 & Yes & No & 8 & -- & -- & 0.583 & 0.634 \\
arousal & 3 & Yes & No & 8 & -- & -- & 0.568 & 0.645 \\
arousal & 1 & Yes & No & 16 & -- & -- & 0.588 & 0.621 \\
arousal & 2 & Yes & No & 16 & -- & -- & 0.599 & 0.603 \\
arousal & 3 & Yes & No & 16 & -- & -- & 0.587 & 0.617 \\
arousal & 1 & Yes & No & 32 & -- & -- & 0.610 & 0.612 \\
arousal & 2 & Yes & No & 32 & -- & -- & 0.584 & 0.610 \\
arousal & 3 & Yes & No & 32 & -- & -- & 0.551 & 0.622 \\
\midrule
valence & 1 & Yes & No & 8 & 0.595 & 0.908 & -- & -- \\
valence & 2 & Yes & No & 8 & 0.538 & 0.909 & -- & -- \\
valence & 3 & Yes & No & 8 & 0.553 & 0.922 & -- & -- \\
valence & 1 & Yes & No & 16 & 0.580 & 0.864 & -- & -- \\
valence & 2 & Yes & No & 16 & 0.547 & 0.900 & -- & -- \\
valence & 3 & Yes & No & 16 & 0.574 & 0.932 & -- & -- \\
valence & 1 & Yes & No & 32 & 0.582 & 0.937 & -- & -- \\
valence & 2 & Yes & No & 32 & 0.609 & 0.874 & -- & -- \\
valence & 3 & Yes & No & 32 & 0.596 & 0.881 & -- & -- \\
\bottomrule
\end{tabular}

\caption{Subtask~2A full experimental results with textual embeddings.}
\label{tab:appendix-full-results-2}
\end{table*}

\begin{table*}
\centering
\scriptsize
\setlength{\tabcolsep}{3pt}
\renewcommand{\arraystretch}{1.1}

\begin{tabular}{l c c c c  C{0.9cm} C{0.9cm}  C{0.9cm} C{0.9cm}}
\toprule
& & & &
& \multicolumn{2}{c}{\textbf{Valence}} 
& \multicolumn{2}{c}{\textbf{Arousal}} \\
\textbf{Target} & \textbf{History length} & \textbf{Text} & \textbf{Semantic cluster} & \textbf{User emb}
& \textbf{{$r$}} & \textbf{MAE}
& \textbf{{$r$}} & \textbf{MAE} \\
\midrule
both & 1 & No & Yes & 8 & 0.531 & 0.948 & 0.580 & 0.645 \\
both & 2 & No & Yes & 8 & 0.585 & 0.874 & 0.631 & 0.614 \\
both & 1 & No & Yes & 16 & 0.530 & 0.925 & 0.553 & 0.660 \\
both & 2 & No & Yes & 16 & 0.542 & 0.915 & 0.581 & 0.616 \\
both & 1 & No & Yes & 32 & 0.597 & 0.863 & 0.583 & 0.613 \\
both & 2 & No & Yes & 32 & 0.602 & 0.862 & 0.602 & 0.622 \\
\midrule
arousal & 1 & No & Yes & 8 & -- & -- & 0.548 & 0.678 \\
arousal & 2 & No & Yes & 8 & -- & -- & 0.591 & 0.636 \\
arousal & 1 & No & Yes & 16 & -- & -- & 0.614 & 0.634 \\
arousal & 2 & No & Yes & 16 & -- & -- & 0.604 & 0.633 \\
arousal & 1 & No & Yes & 32 & -- & -- & 0.620 & 0.629 \\
arousal & 2 & No & Yes & 32 & -- & -- & 0.598 & 0.627 \\
\midrule
valence & 1 & No & Yes & 8 & 0.528 & 0.978 & -- & -- \\
valence & 2 & No & Yes & 8 & 0.523 & 0.953 & -- & -- \\
valence & 1 & No & Yes & 16 & 0.534 & 0.953 & -- & -- \\
valence & 2 & No & Yes & 16 & 0.559 & 0.941 & -- & -- \\
valence & 1 & No & Yes & 32 & 0.601 & 0.859 & -- & -- \\
valence & 2 & No & Yes & 32 & 0.593 & 0.878 & -- & -- \\
\bottomrule
\end{tabular}

\caption{Subtask~2A full experimental results with semantic clusters embeddings.}
\label{tab:appendix-full-results-3}
\end{table*}

\end{document}